\documentclass[11pt,a4paper]{article}
\usepackage{acl2019}
\usepackage{times}
\usepackage{latexsym}
\usepackage{graphicx}
\usepackage{amsmath}
\usepackage{amssymb}  
\PassOptionsToPackage{hyphens}{url}\usepackage{hyperref}

\aclfinalcopy 

\title{The TechQA Dataset}

\author{
Vittorio Castelli$^{1}$ 
\quad
Rishav Chakravarti$^{2}$
\quad
Saswati Dana$^{1}$
\quad
Anthony Ferritto$^{2}$
\quad
Radu Florian$^{1}$ \\
\quad
\textbf{Martin Franz $^{1}$}
\quad
\textbf{Dinesh Garg$^{1}$}
\quad
\textbf{Dinesh Khandelwal$^{1}$}
\quad
\textbf{Scott McCarley$^{1}$}
\quad
\textbf{Mike McCawley$^{3}$} \\
\quad
\textbf{Mohamed Nasr$^{1}$} 
\quad
\textbf{Lin Pan$^{2}$}
\quad
\textbf{Cezar Pendus$^{1}$}
\quad
\textbf{John Pitrelli$^{1}$}
\quad
\textbf{Saurabh Pujar$^{1}$}
\quad
\textbf{Salim Roukos$^{1}$}\\
\quad
\textbf{Andrzej Sakrajda$^{1}$}
\quad
\textbf{Avirup Sil$^{1}$}
\quad
\textbf{Rosario Uceda-Sosa$^{1}$}
\quad
\textbf{Todd Ward$^{1}$}
\quad
\textbf{Rong Zhang$^{1}$}
\\\\
$^{1}$IBM Research AI, \quad $^{2}$IBM Watson, \quad $^{3}$ IBM Finance and Operations\\
\{vittorio, rchakravarti, raduf, franzm, jsmc, mmccawley, mnasr, panl, cpendus, pitrelli,\\ roukos, ansa, avi rosariou, toddward, zhangr\}@us.ibm.com,\\
\{aferritto, saurabh.pujar\}@ibm.com, \{sadana04, garg.dinesh, dikhand1\}@in.ibm.com }

\date{}

\definecolor{darkgreen}{rgb}{0.0, 0.6, 0.13}

\newcommand{\techqa}{{\sc TechQA}}
\newcommand{\TechQA}{{\sc TechQA}}
\newcommand{\hotpotqa}{{\sc HotpotQA}}

\newcommand{\technote}{Technote}
\newcommand{\Technote}{Technote}
\newcommand{\technotes}{Technotes}
\newcommand{\Technotes}{Technotes}

\newcommand{\bertlarge}{BERT$_{\text{LARGE}}$}

\begin{document}
\maketitle
\begin{abstract}

We introduce \techqa, a domain-adaptation question answering dataset for the technical support domain.   The \TechQA\ corpus highlights two real-world issues from the automated customer support domain. First, it contains actual questions posed by users on a technical forum, rather than questions generated specifically for a competition or a task.  Second, it has a real-world size -- 600 training, 310 dev, and 490 evaluation question/answer pairs -- thus reflecting the cost of creating large labeled datasets with actual data.  Consequently, \TechQA\ is meant to stimulate research in domain adaptation rather than being a resource to build QA systems from scratch.
The dataset was obtained by crawling the IBM Developer and IBM DeveloperWorks forums for questions with accepted answers that appear in a published IBM \Technote---a technical document that addresses a specific technical issue.  We also release a collection of the 801,998 publicly available \Technotes\ as of April 4, 2019 as a companion resource that might be used for pretraining, to learn representations of the IT domain language.
\end{abstract}

\begin{figure*}[h!]
    \centering
     \fbox{\includegraphics[width=2\columnwidth]{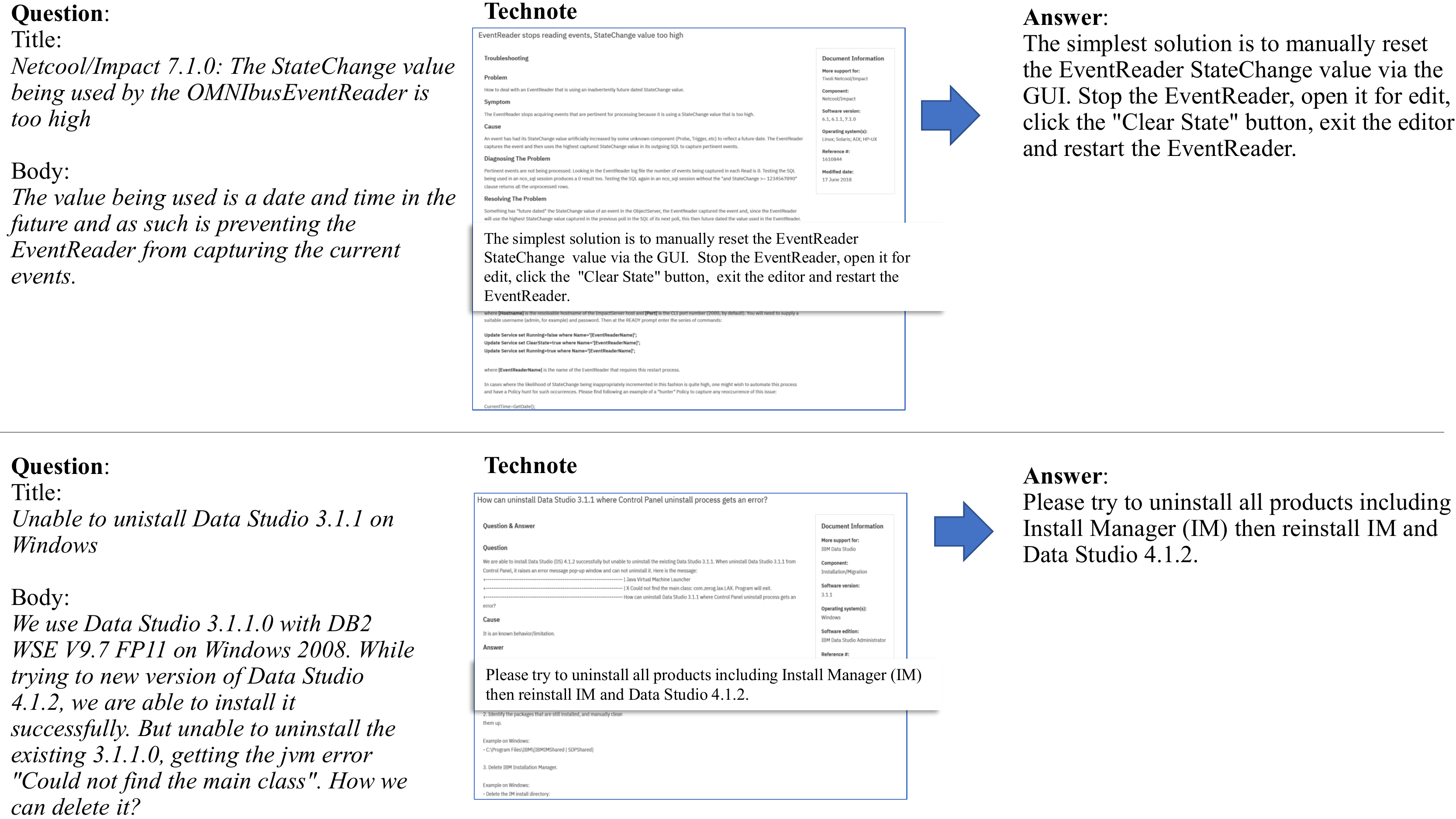}}
    \caption{Examples of questions in the TechQA\ dataset.}
    \label{fig:trainingQuestions}
\end{figure*}

\section{Introduction}
There is a tension between the development of novel capabilities in the early phases of the technology lifecycle, using unlimited data and compute power, and the later development of practical solutions as that technology matures.  The challenges of creating practical solutions are twofold: developing robust, efficient algorithms and curating appropriate training data.  Here we describe the curation and public release of a dataset intended to further those algorithmic advances.

The application domain is IT support, a notable component of the trillion-dollar IT services industry\footnote{\href{https://www.selectusa.gov/software-and-information-technology-services-industry-united-states}{IT Service Report}}.
We created a dataset using publicly available data: questions from technical forums and answers from technical documents.   We have manually selected question-answer pairs that are appropriate for machine reading comprehension techniques, and reserved questions where the answer is distributed across multiple separate spans or multiple documents, and those that require reasoning or substantial real world knowledge for future datasets.   We release 600 questions for training purposes, of which 150 are not answerable from the provided data, as well as 160 answerable and 150 non-answerable questions as development set.  We have reserved 490 questions with similar answerable/non-answerable statistics to the development set as a blind test set.

The purpose of the \TechQA\ dataset is to stimulate transfer learning research from popular question-answering scenarios---driven by large-scale open-domain datasets with short questions and answers---to a use case with involved questions and often long answers. We expect that simple approaches based on tuning models trained on generic datasets will perform poorly on \TechQA, and that systems that are successful at the task need to embody algorithmic advances and novel approaches.  

We are hosting a leaderboard for the \TechQA\ dataset at {\tt ibm.biz/Tech\_QA } where the data---training and development sets, as well as a collection of  more than $800,000$ \Technotes\ published on the internet---is available subject to registration. To maintain the integrity of the test set, the  site provides the tools for authors to submit their system, which we will run on  secure cloud infrastructure.

The rest of the paper is organized as follows.  We briefly review related work in Section~\ref{sec:related}; we then describe the process of collecting the data for \TechQA\ in Section~\ref{sec:collection}, where we detail the automatic filtering, human filtering, annotation guidelines, and annotation procedure.  We present statistics of the dataset in Section~\ref{sec:data_descr}, introduce the associated leaderboard task in Section~\ref{sec:leaderboard} and present baseline results obtained by fine-tuning MRC systems built for Na tural Questions (hence-forth, NQ) \cite{Kwiatkowski2019NaturalQA} and \hotpotqa~\cite{yang2018hotpotqa}  in Section~\ref{sec:baseline}.

\section{Related Work}\label{sec:related}

Recent notable datasets for Machine Reading Comprehension (henceforth, MRC) include the SQuAD 1.1~\cite{Rajpurkar_2016}, SQuAD 2.0~\cite{rajpurkar2018know}, NarrativeQA~\cite{kovcisky2018narrativeqa} and \hotpotqa\ datasets. 
They have stimulated a tremendous amount of research and the associated leaderboards have seen a broad participation across the MRC field.
A common problem of the earlier MRC datasets is observation bias.  Specifically, these datasets contain questions and answers written by annotators who have first read the paragraph that may contain an answer and then wrote the corresponding questions. Hence, the question and the paragraph have substantial lexical overlap.
Additionally, systems trained on SQuAD 1.1 could be easily fooled by the insertion of distractor sentences that should not change the answer as shown in~\cite{jia2017adversarial}. As a result, SQuAD 2.0 added ``unanswerable" questions.
However, large pre-trained language models~\cite{Devlin2018BERTPO, roberta} were able to achieve super-human performance in less than a year on that dataset as well; this suggests that the evidence needed to correctly identify unanswerable questions also are present as specific patterns, such as \textit{antonyms}, in the paragraphs.

Recently, the NQ dataset has been introduced which solves the above problems and constitutes a much harder and realistic benchmark. The questions came from a commercial search engine and were asked by humans who had actual information needs. The answers, which can be long paragraphs, were annotated by human annotators from a Wikipedia page which the user may have selected among the search results. 

Our dataset, \techqa, similarly consists of questions posed in a technical forum by technical users, who had a specific information need, and the answers are in technical documents that another human had linked in the "accepted answer" to the post.  A major structural difference between NQ and \techqa\ is the length of the questions and answers:  NQ questions are about 9 tokens long on average while \techqa\ questions have a title and a body averaging 53 tokens and answers averaging 45 tokens; training answers average above $48$ tokens in \TechQA\ compared to $3.2$ for SQuAD and $2.2$ for \hotpotqa\ (see Section~\ref{sec:data_descr}). 


\hotpotqa\ is a recent multi-hop QA dataset that has questions which necessitate reasoning over text from multiple Wikipedia pages. In addition to answering questions, systems must also extract passages that contain supporting evidence. 

All of the above datasets are supposedly ``open-domain",  as the corpus is Wikipedia. 
There are also datasets for specialized domains. The biomedical QA \cite{tsatsaronis2015overview} dataset contained 29 development questions (arguably too few for training an automated system) and 282 test questions, divided into four categories--`yes/no', factoid, list, and summary.
InsuranceQA \cite{InsuranceQA}, a dataset for the insurance industry, is a corpus for intent detection, rather than for MRC. 

Datasets for specialize domain require effective domain adaptation \cite{wiese2017neural}, because they contain a much smaller number of labeled examples than open-domain datasets like~\cite{bajaj2016ms}.  
Having a limited number of quality labeled examples is a real-world situation: domain experts are much more expensive than crowd-sourcing participants.


\section{\TechQA\ Dataset Collection}\label{sec:collection}
The questions for the \TechQA\ dataset were posed by real users on public forums maintaned and hosted by IBM at the  \href{https://developer.ibm.com/answers/questions}{https://developer.ibm.com/answers/questions} and \href{ https://www.ibm.com/developerworks/community/forums}{https://www.ibm.com/developerworks} sites and have been lightly edited to protect user privacy.  The questions are related to IBM products or other vendors' products running in environments supported by IBM.

Most questions fall into three categories:
i) generic requests for information; ii) requests for information on \emph{how to} perform specific operations;
iii) questions about \emph{causes and solutions} of observed problems.

It appears that the question posers are overwhelmingly \emph{technically knowledgeable people} who have attempted and failed to obtain information before posting on the forum.  The questions are very specific: when describing an issue, the writer typically provides the versions of the affected software products, a description of the operations that yield the error, information about the error including portions of stack traces, and recent changes to the computing environment, such as upgrades, that might have bearing on the problem. 

Questions have both a title and a body. The title is frequently more than just a quick summary of the body, often containing information otherwise not available. In some cases the title and the body are identical, or paraphrases of each other, such as the title:
\emph{``Does anyone have a list of all the versions and fixpacks of the ITCAM Agent for Datapower and from where can I download them?''}	and the corresponding body: \emph{``Does anyone have all the versions of the ITCAM Agent for Datapower which are currently offered for download?''}
Often a question starts in the title and continues into the body.  
We therefore include both the title and body of the question in \TechQA.

\begin{table}
\centering
\small
\begin{tabular}{|l|r|}
\hline
\textbf{Questions} & \textbf{Count}\\\hline
Total retrieved & 276,968\\
With accepted answers & 57,990 \\
With link to Technote in accepted answer & 15,918 \\\hline
\end{tabular}
\caption{Statistics of questions from the forums.  The questions with a Technote in the accepted link were manually annotated by our annotators.}\label{tab:forums}
\end{table}

As outlined in Table~\ref{tab:forums}, a significant fraction of the questions posted in the two forums have answers that were accepted by the original question generator (\emph{accepted answers}).  However, the majority of these accepted answers rely on the question or on fuller forum discourse history and are not good stand-alone candidates for a question/answering dataset.  

For example \emph{``You should be able to debug it -- perhaps the value wasn't populated into that field when the messagebox was called.''} is the accepted answer to the question \emph{``how do I get the value of the dcedFirstName text field to display in my datacap custom verify panel?''}\footnote{This question has been simplified and paraphrased in the interest of space.} Without context, this  answer is uninformative, as are most of the answers in the forums.

About $6\%$ of the accepted answers contain links to one or more \Technotes, documents written and maintained by IBM support personnel that contain information about common questions asked by customers, product upgrade information, and official solutions to well-scoped problems. 
\Technotes\ follow well defined templates: for example,  a troubleshooting \Technote\ has an informative title, a description of the problem, an explanation of the cause, the products, versions, and configurations affected, steps to diagnose the problem, steps to solve the problem, and, if appropriate, temporary workarounds. Metadata in an infobox also describes the components, software version/editions, operating systems, and environments to which the \Technote\ applies, as needed.  

\subsection{Automatic Filtering of Questions}\label{sec:pruning}
The forums were crawled to return only those questions having the following characteristics:
i) the question had an accepted answer; ii) the accepted answer contained a link to a Technote currently published on the web, and  iii) the question was at most 12 sentences long.
 
The last requirement was introduced because most question answering datasets described in Section~\ref{sec:related} contain very short questions; since the goal of the \TechQA\ dataset is to promote domain adaptation, we opt to limit the question length for the \TechQA\ initial release.

The resulting dataset contained 15,918 candidate questions, which were then manually pre-processed as described in the next subsection and subsequently manually annotated by annotators according to the guidelines in Section~\ref{sec:annot_guidelines}

\subsection{Human Annotation}
The step following automatic filtering of questions were performed by six annotators.
Five are professional annotators with substantial experience in NLP annotation.
The sixth is a Linux system administrator.  Four annotators worked full time on the task while the other two, including the system administrator, worked only part time.

With the exception of the system administrator, who also acted in an advisory role, the annotators do not have a technical background.
Crucially, the annotators were not asked to answer technical questions, but to match the content of an accepted answer, provided by a subject matter expert in the forum, with the content of a technical document.
In preparation for \TechQA,  the annotators were trained to annotate \Technotes\ for mention detection according to an unreleased type system we developed for IT technical support. By the time the annotation for \TechQA\ started, our annotators were all sufficiently comfortable with the \Technotes\ terminology an concepts to perform the task.

\subsection{Human Filtering of Questions}
Question filtering consisted of a pass involving the inspection of question titles and bodies only, without considering the answers.  The purpose was to 
 to flag questions that needed manual modification.


Some posts contain multiple questions in the question body.  The prototypical case is a user reporting an error and asking for both cause of and solution to the problem. Also, in some occasions, the title and the body of the question appear to ask for different information as in:
\begin{itemize}
\item title: \emph{``Where can I download the Integration Bus Healthcare Pack''}
  \item body: \emph{``Where can I find information about the Integration Bus Healthcare Pack.''}
\end{itemize}
Such questions were flagged by the annotators and manually split into multiple separate questions each addressing a single information need.  The results of the split were submitted separately to the annotators for manual annotation.  

The annotators also flagged questions that needed to be manually modified as follows:  i) stack traces embedded in questions were reduced by removing irrelevant information; ii) the signoff was removed when it contained a name; iii) product information available from parts of the forum other than the title and text of the questions was worked into the question text,
if this modification was deemed necessary to make the question answerable.   The original questions were disregarded and the modified questions resubmitted for annotation.  

A small fraction of the questions were modified as a results of this and subsequent steps, constituting less than $10\%$ of the released corpus, and most of the changes were very small.

\subsection{Question-Answer Annotation Guidelines}\label{sec:annot_guidelines}
The annotators were instructed to follow the guidelines for question selection and answer span selection outlined below. 

\subsubsection{Question Selection}
Annotators were asked to identify the correct answer in the \Technote\ linked from the forum accepted answer using the question and the accepted answer as guidance.  This strategy enabled non-technical annotators to properly extract answers from the \Technotes.  Using question, accepted answer from the forum and the \technote\ linked in the accepted answer, the annotators were asked to discard questions that had the following characteristics:

i) The accepted answer in the forum is excessively long. Specifically, answers having more than 10 sentences containing information pertaining to solving the problem were discarded. We do this because excessively long accepted answers would impose a substantial cognitive burden on the annotators as they match the content of the \Technote\ with the accepted answer, possibly reducing the quality of the annotation.   It was left to annotators' discretion to retain long accepted answers whenever they felt that the information was clear.

ii) The \Technote\ does not contain an answer to the question.  This happens when the accepted answer points to \Technotes\ that are topical but not essential to the answer.  For example, the answer might state that the product mentioned in the question is an old version that should be updated before addressing the problem and points to a \Technote\ describing the update process.

iii) The answer in the \technote\ is excessively long.  The  guidelines indicated that answers exceeding 10 sentences should be discarded.

iv) The answer consists of multiple separate spans of text. Future releases of the dataset will address domain adaptation for multi-hop question-answering systems.

v) The answer is distributed across multiple \Technotes. 

Additionally, the annotators flagged questions for manual intervention when the information need from the question is ambiguous and only clear upon inspecting the accepted answer.
The authors discarded questions that could not be clarified by means of minimal intervention, and modified and resubmitted for annotation the remaining ones. 

\subsubsection{Answer Span Selection}
The annotators were instructed to select the shortest span that would answer a question under the assumption that the person asking the question is an expert in the field.   
The annotators were also asked to interpret questions literally: if the post asks for the cause of a problem, the answer should not include the solution; conversely, the answer to a post about solving a problem should not contain information about the cause.
When the question title and body appear to ask different questions, the annotator was asked to answer the question in the body.

Text surrounding the actual answer and containing information already provided in the question must not be included in the answer. For example, consider the problem of upgrading a component under Windows$^{\mbox{\textregistered}}$  10 and a Technote that lists the steps for various OS.  The sentence \emph{``These are the steps for Windows$^{\mbox{\textregistered}}$ 10''} should not be part of the selected answer.

Similarly, examples are not deemed to be part of the answer unless they are short and occur in the middle of the answer.

\subsection{Annotation and Adjudication}
Each question that passed the automatic filtering and manual filtering was independently annotated by two annotators, who were not aware of the double annotation procedure.

Questions that were selected by at least one annotator were further manually adjudicated.
The authors reviewed disjoint subsets of the annotator results and were allowed to perform the following operations:
\begin{itemize}
    \item select the answer of one of the annotators when the two annotators disagreed;
    \item reduce the span of the answer, while conforming to the directives listed above;
    \item flag a question as containing multiple questions, when both annotators failed to recognize it;
    \item shorten the question, mostly by removing parts of stack traces (a process that could be easily automated);
    \item occasionally reject the answer---by-and-large when one of the annotators had already rejected the answer.
\end{itemize}

The two authors also independently annotated 100 answerable questions; the resulting inter-annotator agreement F1 is $76.3\%$  and the exact match rate is $61\%$.

The resulting set of question/answer pairs contains  slightly  more  than  1000  items.   In  future versions  of  the  \TechQA,  we  plan  to  relax  many of these annotator constraints to promote research addressing  a  broader  spectrum  of  technical  support problems.
\section{\TechQA\ Dataset Characteristics}\label{sec:data_descr}
The \TechQA\ dataset consists of a training set, a development set, a test set, and a small validation set.
The training set contains 450 answerable questions and 150 non-answerable questions, the development set consists of 160 answerable and 150 non-answerable questions, and the test set consists of 490 questions with similar answerable vs. non-answerable proportions as the development set.  The validation set consists of the first 20 entries of the development set and is used in the leaderboard described in Section~\ref{sec:leaderboard}.
We also provide the full collection of the unique $801,998$ \Technotes\ that were available on the web as of  April 4, 2019.

The dataset is designed for machine reading comprehension, rather than for open-domain question answering. Specifically, instead of requiring users to search the Technote collection to find one containing the answer, we provide for each question a candidate list of 50 Technote IDs.  Systems should analyze only the 50 \Technotes\ associated with the question.  A question is answerable if the annotators found an answer in one of these 50 \Technotes, and is unanswerable otherwise. Systems can access the entire \Technote\ collection but only answers from the 50 \Technotes\ associated with each questions will be scored.

The 50 \technotes\ were obtained by issuing a query to an instance of Elasticsearch\footnote{\href{https://www.elastic.co/products/elasticsearch}{https://www.elastic.co/products/elasticsearch}} that indexes the $801,998$ \Technotes. This query consisted of the concatenation of the question title and question text.  Consequently, the retrieved \Technotes\ are expected to contain at least some of the low-frequency terms in the question.

\begin{figure*}
    \centering
    \includegraphics[width=3.5in]{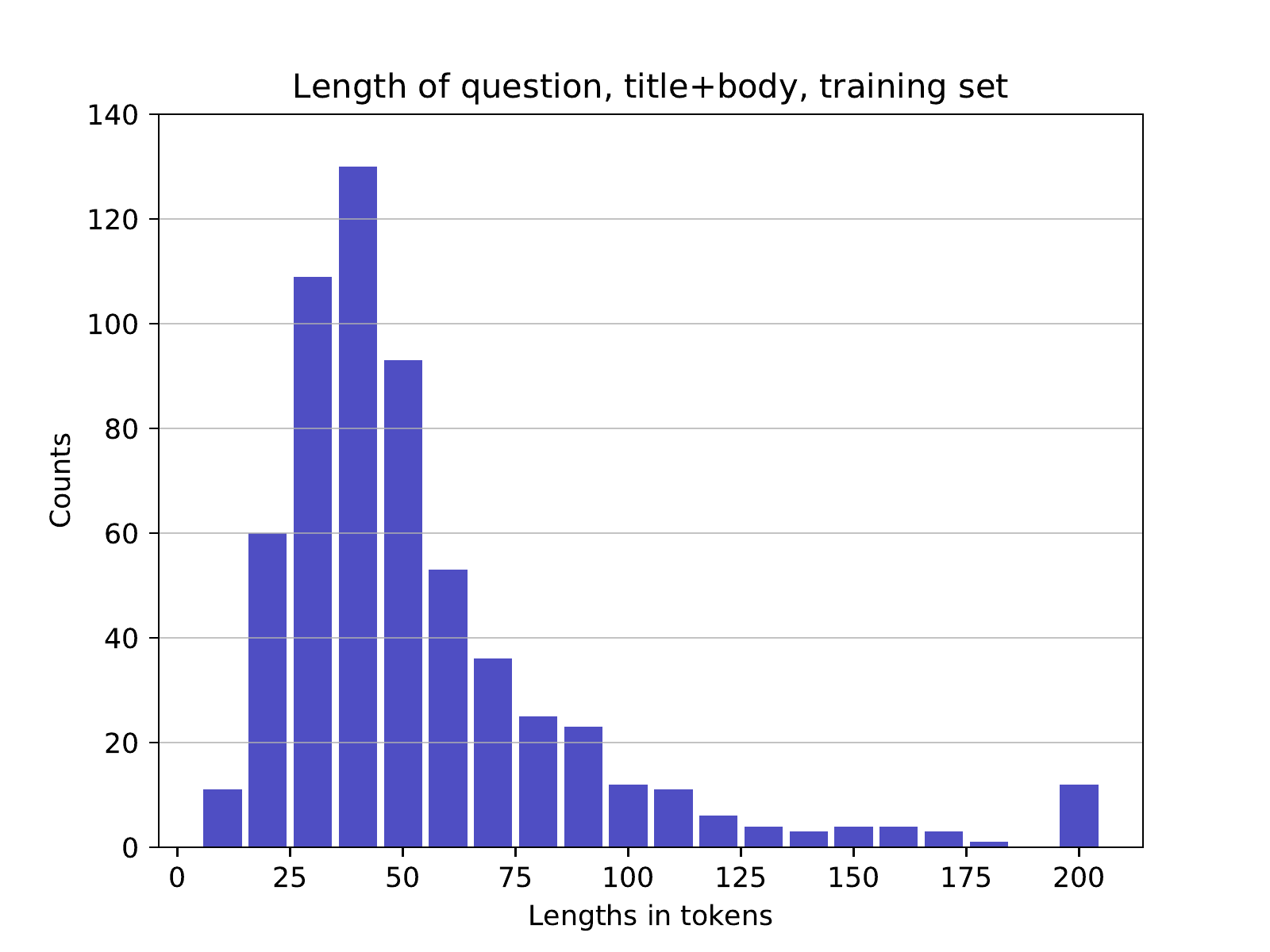}\includegraphics[width=3.5in]{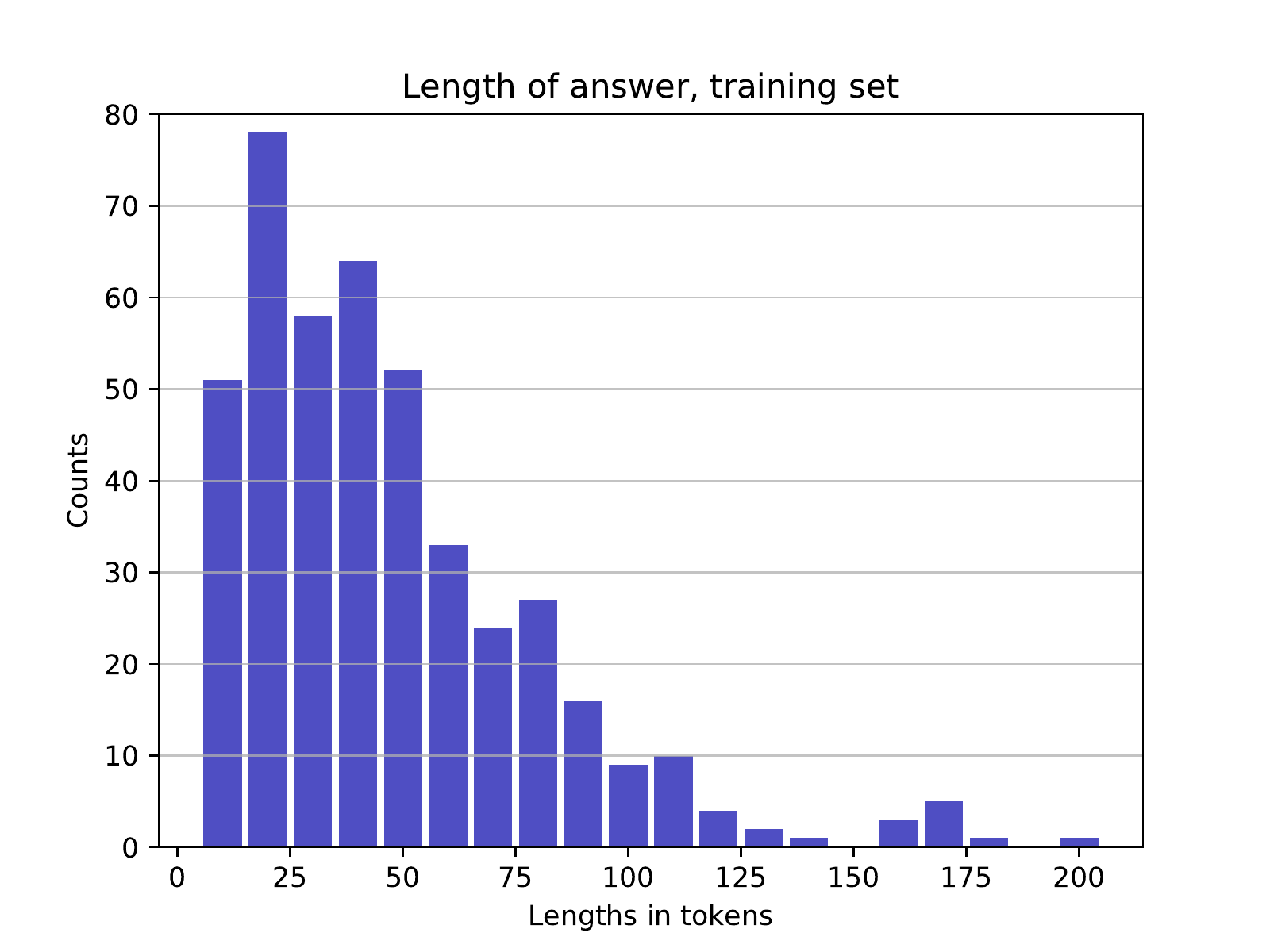}
    \caption{Number of white space separated tokens in training questions (title plus body.) and answers (for answerable questions only). The bin at 200 also contains all questions longer than 200 tokens.}
    \label{fig:trainingQA}
\end{figure*}

\begin{figure*}
    \centering
    \includegraphics[width=3.5in]{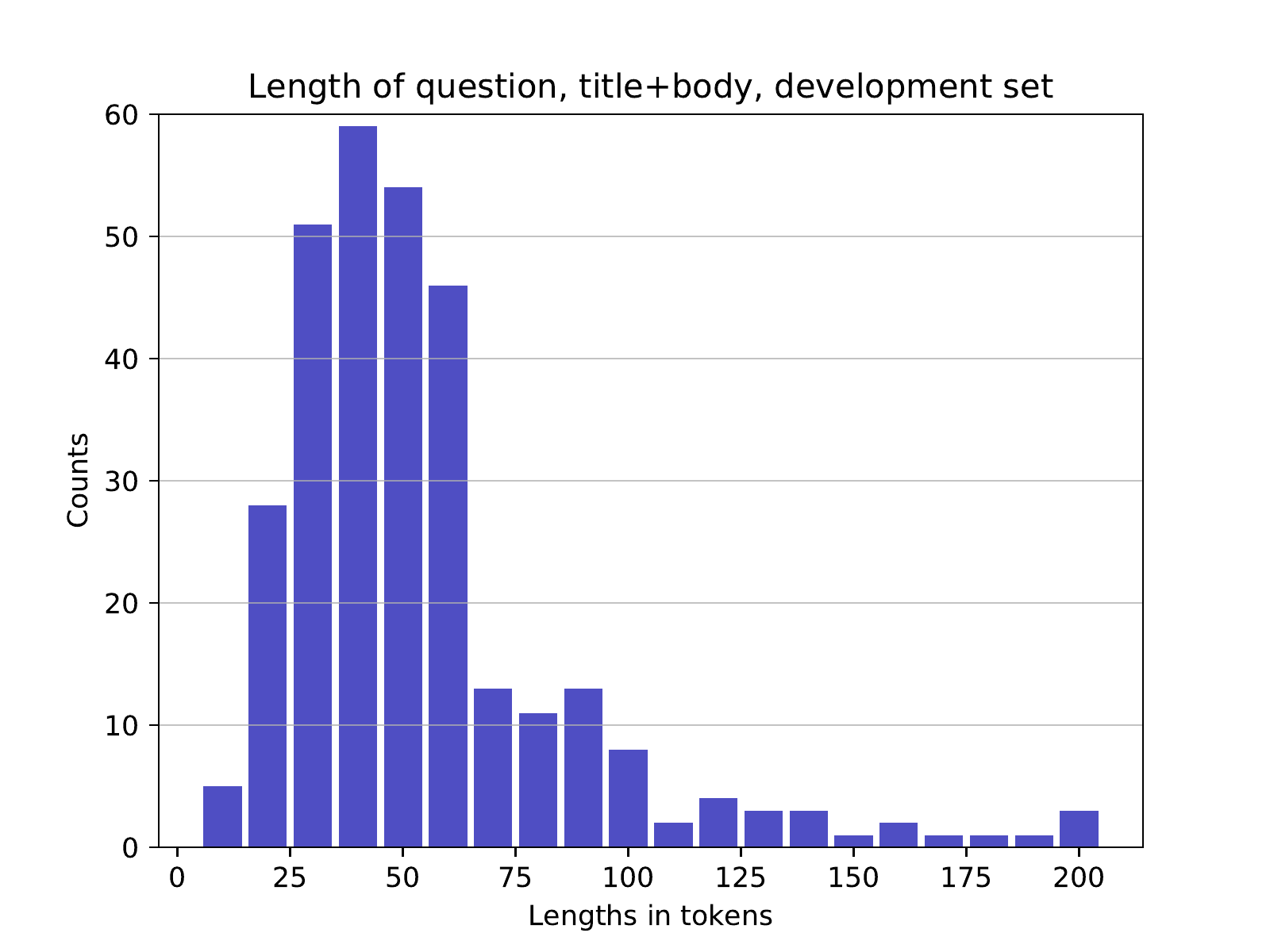}\includegraphics[width=3.5in]{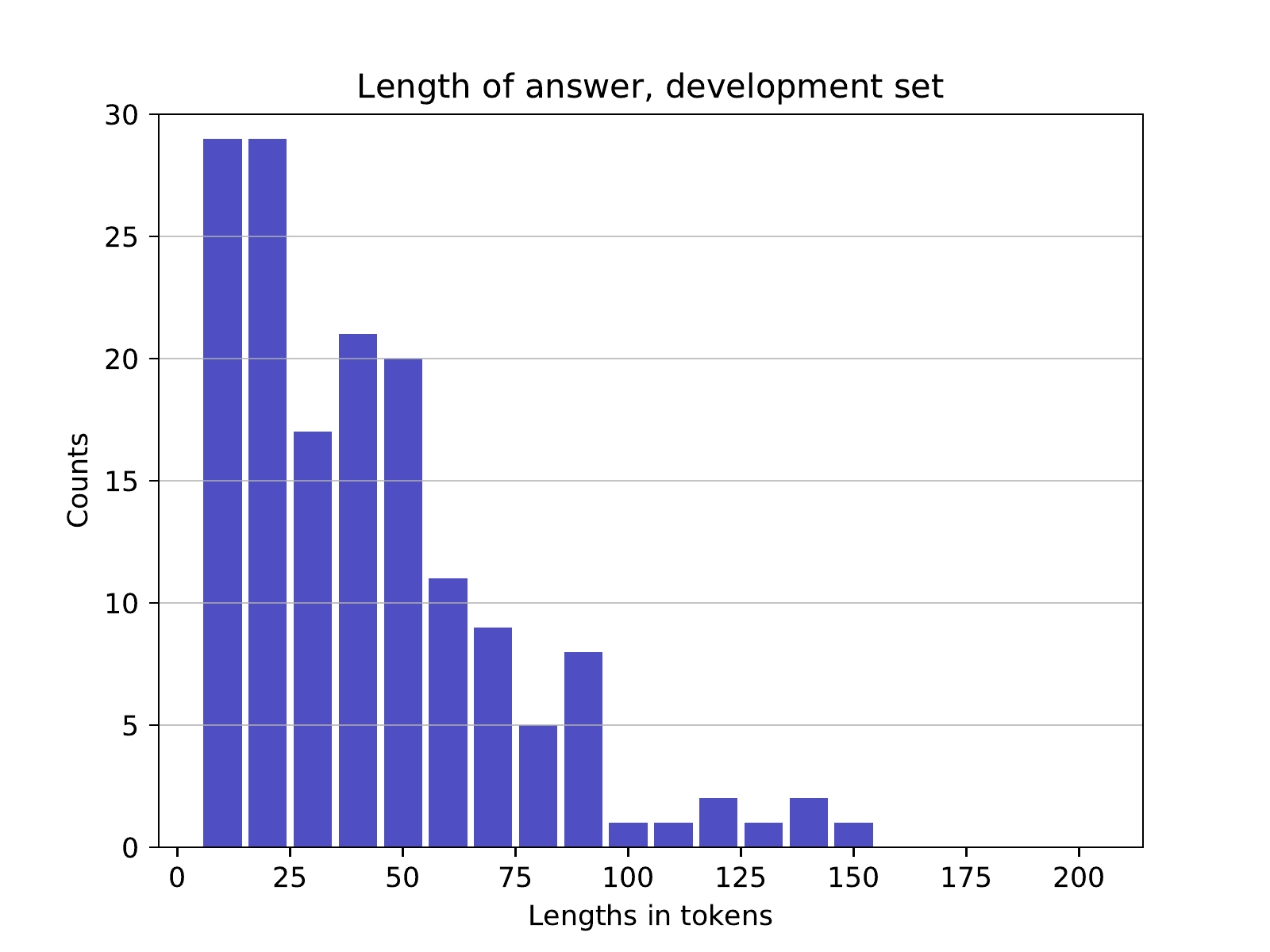}
    \caption{Number of white space separated tokens in devtest questions (title plus body) and answers (for answerable questions only). }
    \label{fig:devQA}
\end{figure*}

\TechQA\ questions and answers are substantially longer than those found in common datasets. Table~\ref{tab:tok_stats} compares statistics of training and development sets questions and answers of \TechQA\ to those of {SQuAD} 2.0 and {\hotpotqa}, in white-space-separated tokens.   Figures~\ref{fig:trainingQA} and \ref{fig:devQA} depict the length distributions for questions text, title plus text, and answers for training and devset, respectively.  Most questions have a length between 10 and 75 tokens, but the dataset exhibits a long tail, reflecting the fact that questions with a substantial amount of detailed information are relatively common.  Most answers are between 1 and 100 tokens long, and the distribution has a long tail.  

These structural differences should pose interesting challenges to domain adaptation.

\begin{table*}
\centering
\begin{tabular}{|l|l|r|r|r|r|l|r|r|r|r|r|}
\hline
\multicolumn{2}{|l}{~}&\multicolumn{4}{|c|}{\bf Question length in tokens}&&\multicolumn{4}{|c|}{\bf Answer length in tokens}\\
\hline
{\bf Dataset}     & {\bf split} & min & mean & max & std &{\bf split}& min & mean & max & std\\
\hline
{SQuAD 2.0} & training & 1 & 9.9& 40& 3.4 &training& 1& 3.2&  43 & 3.4 \\
& devset   & 3 & 10.0 & 31& 3.45 &devset& 1 & 3.1&  29 & 3.1\\
\hline
\hotpotqa 
& training& 3& 17.8 & 108 & 9.5 &training& 1 & 2.2& 89 & 1.8\\
& devset  & 6 & 15.7& 46  & 5.5 &devset& 1& 2.5& 29& 1.8 \\
\hline
\TechQA
& training & 8 & 52.1 & 259 & 31.6 &training& 1 & 48.1 & 302 & 37.8 \\
& devset   & 10 & 53.1 & 194 & 30.4 &devset& 1  & 41.2 & 137 & 27.7 \\
\hline
\end{tabular}
\caption{Statistics of the question and answer lengths in white-space-separated tokens for SQuAD 2.0, \hotpotqa\ and {\TechQA}.}
\label{tab:tok_stats}
\end{table*}                                 
                            

Finally, the question and answers contain numerous technical terms which are likely not part of the vocabulary of most contextual language models.   One of the reasons for including the whole \Technotes\ corpus is to provide data for enhancing the language models by appropriately enlarging the vocabulary to include technical support terms.
\section{Leaderboard task}\label{sec:leaderboard}
The dataset is available by registering to the leaderboard at {\tt ibm.biz/Tech\_QA}.
Once registered, a user has access to the data and to means for submitting systems for evaluation against the blind test set.  As with other leaderboards, this approach will help maintain the integrity of the evaluation set. 

A system to be submitted for evaluation must be packaged as a Docker image, containing all the needed components.  The image will be run in isolation from the network and systems will not be able to download data while running in the leaderboard environment. The systems will be able to read the data from a specific read-only input directory and to write results to a specific output directory.  These are the only directories mounted into the image that are visible to the system.  Detailed instructions on how to package the system are available from the leaderboard site.
Crucially, we ask that systems submitted to the leaderboard do not use information from the
\url{ developer.ibm.com} and \url{www.ibm.com/developerworks} web sites except for the data provided with the dataset.

Submitted systems will run on a machine with 128 GB of memory and two 16G V100 GPUs, with 64 GB local disk space available for temporary files or logs. 

Upon submission, the system will be run against the validation set provided with the data distribution. This contains the first 20 questions from the devset.  Upon completion, the results of the validation run are available from the user's personal dashboard, for comparison with results computed by the user.  A user satisfied with the validation run can submit the system to be run against the 490 evaluation questions. Runs will be limited to 24 hours, after which they will be terminated and the submission will be in an error state in the dashboard. Once the run is complete, it is added to the dashboard.  

The user can monitor the progress of each submission from the dashboard, and cancel the submission at any point previous to completion of the evaluation run. Once the evaluation run succeeds, the results are posted on the leaderboard. A user is prevented from submitting a new system for a week starting from the date of the most recent submission, as it appears on the leaderboard.
The dashboard provides means for anonymizing and de-anonymizing a successful submission (for example, for paper review purposes).  An anonymized submission retains the name of the system provided by the user, but hides the user's affiliation as well as the optional link to a paper.

Systems are required to analyze the 50 documents associated with each question, and produce 5 candidate answers.  Each answer consists of a document ID,  a pair of character offsets from the beginning of the detagged text of the \Technote\, and a score. The score is compared with a threshold provided by the system and associated with the entire run over all the documents.   Systems must return scores lower than the threshold to indicate that no answer exists in the \Technote; however, they also must indicate the best span extracted from the document: this is used to compute the two ancillary metrics described below.

The evaluation score computed for the leaderboard is a zero-one value for a question/document pair with score below the threshold, and character-overlap F1 for a question/document pair with sore greater than or equal to the threshold.  

The main metric, denoted as F1 on the leaderboard, is the macro average of the evaluation scores computed on the first of the five answers provided by the system in response to each question.

The leaderboard displays two ancillary metrics.  HA\_F1@1 is the macro average of the evaluation scores computed on the first of the five answers and averaged over the answerable question subset of the evaluation set.  This metric should be compared to the inter-annotator agreement of $76.3$ reported in Section~\ref{sec:collection}.
HA\_F1@5 consists of computing the evaluation score for each of the 5 answers, selecting the maximum, and computing the macro average over all answerable questions.

The leaderboard also reports BEST\_F1, the value of the F1 metric corresponding to the optimal choice of the threshold.

The leaderboard is seeded with the baseline results described in the next section.
\section{Baseline Results}\label{sec:baseline}

\begin{table*}[ht!]
\centering
\begin{tabular}{lrrrr}
\hline
\textbf{Systems}  & \makebox[.7in][r]{\textbf{F1}}	& \makebox[.7in][r]{\textbf{HA\_F1$@$1}}	& \makebox[.7in][r]{\textbf{HA\_F1$@$5}}	& \textbf{BEST\_F1}  \\
\hline
\hline
SQuAD 2.0 $-$ FT & 1.67$^{\;\;\,}$	& 3.25	& 4.51	& 48.39 \\
\hline
SQuAD 2.0 $+$ FT & 54.05$^{\ast}\,$	& 22.01	& 35.50	& 54.05 \\
\hline
NQ $-$ FT  & 2.74$^{\;\;\,}$	& 5.32	& 9.07	&48.39\\
\hline
NQ $+$ FT & \textbf{55.31}$^{\ast}$& \textbf{34.69}	& 50.52	& \textbf{55.31} \\
\hline
{TAP\_v0.1} & \textbf{51.36}$^{\;\;}$ & 16.39 & \textbf{57.49} & 52.67 \\ 
\hline
\end{tabular}
\caption{Our baseline systems on the dev set. Here, `$-$FT' indicates no fine-tuning and we use a pre-trained SQuAD 2.0 and NQ models, while `$+$FT' indicates further fine-tuning using the \TechQA\ corpus. Entries marked with `$\ast$' use a threshold tuned on the development set using the F1 metric; hence, F1 equals {BEST\_F1}.}
\label{tab:baseline-results}
\end{table*}

Table~\ref{tab:baseline-results} show the results of our baseline systems.  These are a model trained on SQuAD 2.0,  a model trained on NQ, and the TAP system submitted to the \hotpotqa\ leaderboard.  The SQuAD and NQ models are shown with and without fine tuning on the \TechQA\ dataset.

\subsection{SQuAD and NQ}

We use a model pre-trained on SQuAD 2.0 and one pretrained on NQ which is SOTA on the short answer leaderboard.
Both models start from the \bertlarge{} (whole word masking) language model \cite{Devlin2018BERTPO} which 
takes a maximum 512 input word piece token sequence $\mathbf{X} = [x_1, x_2,\ldots,x_T]$ and uses a $L=24$ 
layer Transformer \cite{Vaswani_2017} network (with 16 attention heads and 1024 embedding dimensions) to output a sequence of contextualized token representations $\mathbf{H}^L = [\mathbf{h}_1^L, \mathbf{h}_2^L,\ldots,\mathbf{h}_T^L]$.  The input sequence $\mathbf{X}$ is marked up with special tokens ({\tt[CLS]} and {\tt[SEP]}) to facilitate learning text boundaries and classification tasks. Specifically, for the QA or MRC task, $\mathbf{X}$ consists of a {\tt[CLS]} token, followed by the query tokens followed by a {\tt[SEP]} token, followed by the (potential) answer containing document span tokens, followed by a final {\tt[SEP]} token.

For our SQuAD 2.0 model, we follow \cite{Devlin2018BERTPO} by adding two fully connected feed forward layers followed by a softmax 
for answer extraction: $\boldsymbol{\ell}_b = softmax(\mathbf{W}_1 \mathbf{H}^L)$ and $\boldsymbol{\ell}_e = softmax(\mathbf{W}_2 \mathbf{H}^L)$, where $\mathbf{W}_1$, $\mathbf{W}_2 \in \mathbb{R}^{1\times 1024}$. $\boldsymbol{\ell}_b^t$ and $\boldsymbol{\ell}_e^t$ denote the probability of the $t^{th}$ token in the sequence being the answer beginning and end, respectively.

For the NQ model we start from the SQuAD 2.0 model and follow \cite{alberti2019bert} to add a third layer for target type 
prediction: $\boldsymbol{\ell}_a = softmax(\mathbf{W}_3 \mathbf{h}_{[CLS]}^L)$, $\mathbf{W}_3 \in \mathbb{R}^{5\times 1024}$ \footnote{The $5$ dimension in $\mathbf{W}_3$ corresponds to the $5$ answer target types: short answer, long answer, yes, no, and no answer.},
and $\mathbf{h}_{[CLS]}^L \in \mathbb{R}^{1024}$. 
We employ the  hyper-parameter tuning and training strategies detailed in \cite{pan2019frustratingly}.
Despite both NQ and SQuAD 2.0 being well aligned with the \techqa\  task (including the presence of un-answerable questions), 
we see that plain decoding with pre-trained SQuAD 2.0 and NQ model is not helpful on the \techqa\ dataset; further demanding the need to fine-tune on the provided target domain training data. 

When fine tuning further for the \techqa\ dataset, we start from the
SQuAD and NQ models
and continue training drawing on hyper-parameters from \cite{pan2019frustratingly} with the following adjustments:
\begin{itemize}
    \item The query title and body are concatenated (with a dividing {\tt[SEP]} token), truncating as needed to limit both to 110 total word pieces.
    \item The document title is included at the beginning of each document span (with a dividing {\tt[SEP]} token).
    \item The negative span sub-sampling rates that compensate for class imbalance are changed to 0.1 and 0.15 when the question is or is not answerable, respectively, from their NQ values of 0.01 and 0.04.
\end{itemize}

We see that fine-tuning actually helps (6.9 F1 points improvement) over just decoding with a pre-trained NQ model. We remind the readers that this 
architecture and training strategy 
\cite{pan2019frustratingly} is the current SOTA system on the NQ short answer leaderboard\footnote{At the time of writing of the paper.} and even outperforms single-human performance on that dataset.


\subsection{The TAP Baseline System}
Our second baseline is adopted from the TAP submission on the \hotpotqa\ leaderboard\footnote{https://hotpotqa.github.io/}. We modified the TAP architecture to adapt it here. 
We call this adapted baseline system {\texttt{Technical Answer Prediction (TAP)}}. 
TAP is a cascade-style architecture comprising of two modules - (1) {\em document-ranker\ (aka ranker)}, and (2) {\em span-selector}. At inference time, the ranker generates a ranked list of the 50 documents (along with scores $s_1, s_2, \ldots, s_{50}$) that are supplied with any test question. Based on a pre-decided threshold value $\lambda$, the question is classified into ``answerable", if $\lambda \le \underset{i=1 \rightarrow 50}{\max}\;\{s_i\}$, and ``unanswerable", otherwise. If the question is predicted to be ``answerable"; its rank-1 predicted document (along with question string) is supplied to the span-selector, which in turn predicts the answer span from within the supplied document.


During training, the ranker and the span-selector are trained independently. Both the ranker as well as the span-selector are built on top of the pre-trained BERT-base model, where we supply the token sequence \texttt{[CLS, QUE, SEP, DOC, SEP]} as input. The \texttt{QUE} corresponds to a maximum of
the first 80 tokens (padded and trimmed as necessary) of the question string, and \texttt{DOC} corresponds to a minimum of the first 429 tokens from the document (padded and trimmed as necessary) so as to make a 
512 length token sequence. To enhance the coverage of the document beyond 429 tokens, we actually use two such token sequences each having 429 tokens of the documents (thereby, covering a minimum of 858 tokens), and each supplied to its respective BERT. In the case of ranker, the \texttt{CLS} token output from each copy of the BERT are concatenated and supplied to a linear layer which assigns a score value $s_i$. In the case of span-selector, the 858 output tokens corresponding to the document are passed through a linear layer to reduce dimension from 768 to 512 followed by single layer of Transformer Encoder~\cite{Vaswani_2017}. Each output token from the Encoder layer is fed into a linear layer to get two probability scores - {\em{start probability}}, and {\em{end probability}}. The training loss for both the ranker as well as the span-selector is the cross entropy loss. The threshold $\lambda$ is chosen by tuning it to maximize the ranker's accuracy on the development set.
The number of training epochs used were 2 and 4 for the ranker and the span-selector, respectively.
\section{Discussion and Future Work}\label{sec:conclusions}
We have introduced \TechQA, a question-answering dataset for the IT technical support domain.  

  The overall size of the released data (600 training questions) is in line with real-world scenarios, where the high cost of domain expert time limits the amount of quality data that can reasonably be collected.  Thus, the dataset is meant to stimulate research in domain adaptation, in addition to developing algorithms for longer questions and answers than the current leaderboards.

We have created a leaderboard to evaluate systems against a blind dataset of 490 questions with a ratio of answerable to unanswerable questions similar to that of the development set.    The leaderboard ranks submissions according to a metric consisting of the character overlap F1 measure for answerable questions and the zero-one metric for non-answerable questions.  In addition, the leaderboard reports the F1 at the top result and the F1 for the top 5 results computed over the answerable test questions.

\TechQA\ is a challenging dataset for models developed for existing open-domain MRC systems. Their out-of-the box performance is very low, especially considering that a system that declares every question as unanswerable achieves F1=$48.4\%$ on the development set.  The obvious approach of fine-tuning these models using the \TechQA\ training set yields systems that barely beat the baseline. 

The initial version of the dataset was created by selecting questions and answers that are relevant to the IT technical support domain but at the same time do not diverge excessively from the spirit of other existing MRC datasets. We consider \TechQA\ to be an important stepping stone on which to build future data collections and leaderboards.

We plan on releasing questions with answers in a broader and more diverse collection that will include documents with a less formulaic structure than the \technotes.  We will also relax the length limitations to include questions rich in details, and answers that include complex procedures; in the same spirit, we will allow answers consisting of multiple spans from a single document. 

Many answers cannot be obtained by extracting portions of a documents based on language alone: in many cases, domain knowledge is needed and often a question cannot be answered from the data collection without a reasoning step. We envision a roadmap where future releases of \TechQA\ will require synergy between multiple AI disciplines, from deep-learning based MRC to reasoning, knowledge base acquisition, and causality detection.

\section*{Acknowledgments}
We would like thank our annotators: Abraham Mathews (IBM), Kat Harkavy,  Irina Paegelow, Daniele Rosso,  Chie Ugumori and Eva Maria Wolfe (ManpowerGroup Associates) for their dedication to the project and their relentless  annotation effort.
\bibliographystyle{acl_natbib}
\bibliography{gaama}

\end{document}